\documentclass[letterpaper]{article}

\usepackage[preprint]{colm2026_conference}

\usepackage{microtype}
\usepackage{hyperref}
\usepackage{url}
\usepackage{booktabs}

\usepackage{graphicx}
\usepackage{amsmath}
\usepackage{amssymb}
\usepackage{mathtools}
\usepackage{amsthm}

\usepackage{multirow}
\usepackage{makecell}
\usepackage{xcolor}
\usepackage{enumitem}
\usepackage[ruled,vlined]{algorithm2e}

\usepackage{seqsplit}  
\usepackage{tcolorbox} 
\newtcolorbox{codebox}[1]{
  colback=white,
  colframe=black,
  colbacktitle=white!90!gray,
  coltitle=black,
  fonttitle=\bfseries,
  title={#1},
  arc=0mm,
  boxrule=0.8pt,
  left=2pt, right=2pt, top=2pt, bottom=2pt
}

\usepackage{lineno}

\definecolor{darkblue}{rgb}{0, 0, 0.5}
\hypersetup{colorlinks=true, citecolor=darkblue, linkcolor=darkblue, urlcolor=darkblue}

\title{Reinforced Reasoning for End-to-End \\ Retrosynthetic Planning}

\author{
  Chenyang Zuo\textsuperscript{1,2} \quad
  Siqi Fan\textsuperscript{1} \quad
  Yizhen Luo\textsuperscript{1} \quad
  Zaiqing Nie\textsuperscript{1,2}\thanks{Corresponding author. For any discussions, please email \texttt{zuocy22@mails.tsinghua.edu.cn}, \texttt{fansiqi@air.tsinghua.edu.cn}.} \\
  \\
  \textsuperscript{1}Institute for AI Industry Research (AIR), Tsinghua University \quad
  \textsuperscript{2}PharMolix Inc.
}

\begin{document}

\ifcolmsubmission
\linenumbers
\fi

\maketitle

\begin{figure}[h]
\begin{center}
\includegraphics[width=1.0\linewidth]{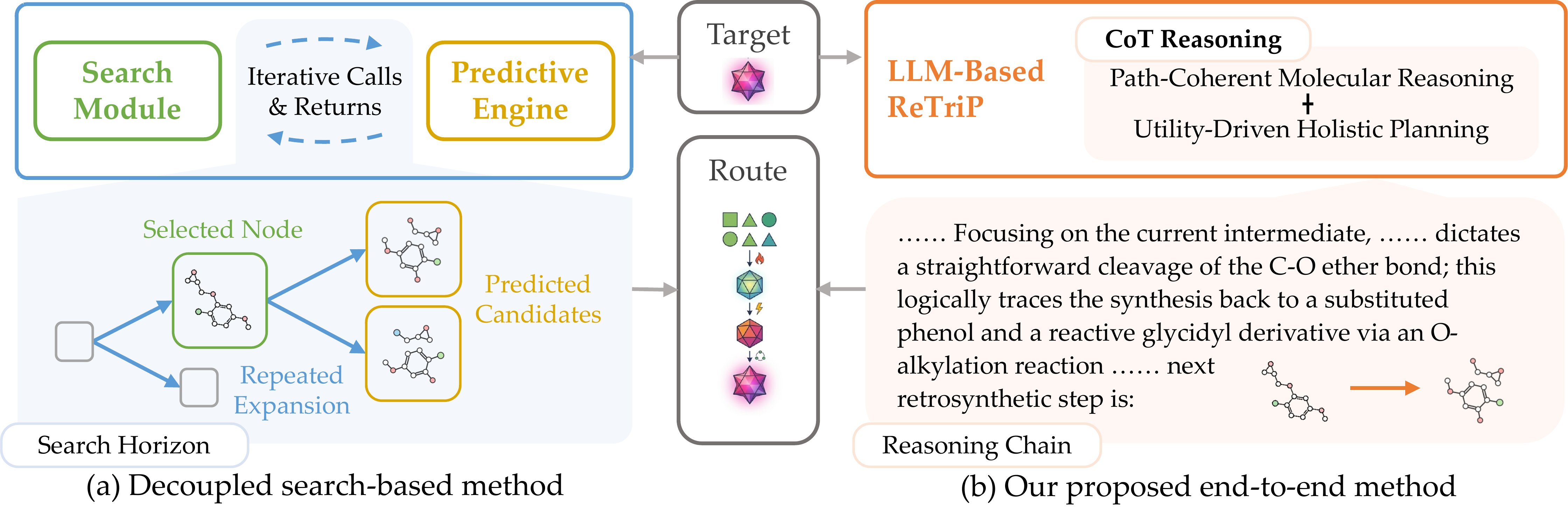}
\caption{\textbf{Comparison of retrosynthetic paradigms.} (a) Decoupled search-based method; (b) Our unified ReTriP framework.}
\label{fig:paradigm_single}
\end{center}
\end{figure}

\begin{abstract}
Retrosynthetic planning is a fundamental task in organic chemistry, yet remains challenging due to its combinatorial complexity. To address this, conventional approaches typically rely on hybrid frameworks that combine single-step predictions with external search heuristics, inevitably fracturing the logical coherence between local molecular transformations and global planning objectives. To bridge this gap and embed sophisticated strategic foresight directly into the model's chemical reasoning, we introduce ReTriP, an end-to-end generative framework that reformulates retrosynthesis as a direct Chain-of-Thought reasoning task. We establish a path-coherent molecular representation and employ a progressive training curriculum that transitions from reasoning distillation to reinforcement learning with verifiable rewards, effectively aligning stepwise generation with practical route utility. Empirical evaluation on RetroBench demonstrates that ReTriP achieves state-of-the-art performance, exhibiting superior robustness in long-horizon planning compared to hybrid baselines.
\end{abstract}

\section{Introduction}

Retrosynthetic planning is a fundamental problem in organic chemistry, critical for applications ranging from drug discovery to chemical biology \citep{corey1991logic, gao2020synthesizability}. This task identifies synthetic routes by recursively decomposing target molecules into available precursors \citep{corey1967general, corey1985computer}. However, the exponential expansion of potential reaction pathways as route length increases creates a vast combinatorial search space \citep{szymkuc2016computer}. Navigating this space manually is dauntingly complex and time-consuming, necessitating automated long-horizon planning \citep{coley2019robotic}.

Mainstream machine learning frameworks typically address multi-step retrosynthesis by combining predictive engines with external search algorithms (Figure~\ref{fig:paradigm_single}a) \citep{zhong2024recent, comprehensiveSurvey}. In this hybrid paradigm, predictive engines generally operate as single-step expansion modules that propose immediate precursors for a given molecule \citep{segler2017neural, liu2017retrosynthetic, zhong2023retrosynthesis}, while search modules employ layered exploration strategies to assemble these local decompositions into complete synthetic routes \citep{segler2018planning, chen2020retro, yu2024double}. Recently, Large Language Models (LLMs) have been integrated into this pipeline as modular components, serving as single-step reasoners \citep{zhang2025reasoningdrivenretrosynthesispredictionlarge, li2025retroexpertcollaborativereasoninginterpretable}, agentic controllers for single-step tools \citep{bran2023augmenting, liu2025retror}, or evolutionary path refiners \citep{wang2025llmaugmented}. However, the decoupling of expansion and exploration inevitably fragments chemical knowledge, disrupting the logical coherence between local transformations and the global retrosynthetic strategy. This myopic approach to decision-making undermines both the accuracy and robustness of the generated pathways. Bridging this gap necessitates a unified paradigm that embeds multi-step causal dependencies directly within its generation process, ensuring that every predicted step is strategically grounded.

To realize this unified design, we propose ReTriP (\textbf{Re}inforced \textbf{Re}asoning for \textbf{Re}trosynthetic \textbf{P}lanning), an end-to-end framework that formulates retrosynthetic planning as an explicit reasoning task (Figure~\ref{fig:paradigm_single}b). Leveraging LLM-driven Chain-of-Thought (CoT) reasoning, ReTriP effectively captures long-range chemical dependencies while reconciling complex trade-offs among diverse synthetic constraints. Developing such a framework, however, presents non-trivial challenges. A primary bottleneck stems from standard SMILES notation \citep{weininger1988smiles}. By representing molecular species as isolated entities, this notation inherently causes inconsistent atom indexing across consecutive reaction steps \citep{zhong2022root, tu2022permutation}. This inconsistency obscures underlying topological transformations and disrupts route-level fragment traceability \citep{schwaller2021extraction, liu2023fusionretro}. Consequently, establishing a path-coherent molecular representation becomes a strict prerequisite for enabling chemically grounded, route-level reasoning. Furthermore, achieving practical synthetic utility requires overcoming the objective misalignment inherent in standard imitation learning, which struggles to generalize and fails to guarantee efficiency metrics such as route parsimony \citep{lin2020automatic, kim2021self, shee2025directmultistep}. Addressing this limitation necessitates integrating knowledge-grounded linguistic reasoning with a goal-oriented optimization paradigm, leveraging fine-grained feedback to explicitly align the model's stepwise reasoning with long-term synthetic viability.

To establish representational coherence, ReTriP leverages informative target notations as model input and goes beyond the localized R-SMILES alignment \citep{zhong2022root} to maintain continuous structural correspondence across long synthetic routes. To achieve utility-oriented planning, the framework adopts a progressive curriculum that advances from CoT distillation to Reinforcement Learning with Verifiable Rewards (RLVR). This synergy between fragment-traceable molecular representations and goal-oriented optimization empowers ReTriP with path-coherent reasoning and utility-driven holistic planning. Experimental results on RetroBench demonstrate that ReTriP achieves a Top-1 accuracy of 45.5\%, outperforming the previous state-of-the-art (SOTA) \citep{kang2025retrointext} by 3.4 percentage points. Notably, our framework exhibits superior robustness in long-horizon planning (depth $\ge$ 5). Our contributions are summarized as follows:

\begin{itemize}
  \item \textbf{Unified retrosynthetic paradigm:} We internalize multi-step retrosynthetic logic into a standalone CoT process for end-to-end planning, effectively eliminating the reliance on decoupled search heuristics.
  \item \textbf{Methodological innovation:} We develop a progressive optimization curriculum that integrates path-coherent molecular representations with goal-oriented RLVR, bridging the gap between local reaction logic and long-term synthetic utility.
  \item \textbf{SOTA performance:} We establish a new SOTA on RetroBench, surpassing traditional hybrid architectures in both overall accuracy and long-horizon robustness.
\end{itemize}

\section{Related work}

\textbf{Search-based retrosynthetic planning.} Traditionally, retrosynthesis is addressed through a decoupled framework that pairs a single-step predictive engine with an external heuristic search algorithm. The single-step models are broadly categorized into template-based \citep{coley2017computer, segler2017neural, gln, Chen2021DeepRR, Xie2023RetrosynthesisPW}, semi-template-based \citep{shi2020graph, somnath2021learning, zhong2023retrosynthesis}, and template-free methods \citep{liu2017retrosynthetic, zheng2019predicting, karpov2019transformer, schwaller2020predicting, sacha2021molecule} based on their reliance on expert-defined rules. These engines serve as expansion modules, orchestrated by search algorithms such as MCTS \citep{segler2018planning, hong_retrosynthetic_2023}, Depth-First Proof-Number search \citep{dfpn-e}, or A*-like heuristics \citep{chen2020retro, zhao2024efficient} to navigate the combinatorial search space. Given the heavy computational overhead of this hybrid design, recent research focuses on optimizing search efficiency from two primary perspectives. The first involves aligning local single-step predictions with global planning objectives through reinforcement learning (RL) \citep{kim2021self, yu2022grasp, liu2023retrosynthetic} or in-context learning \citep{liu2023fusionretro, kang2025retrointext}. The second perspective focuses on advancing the search mechanism by refining the pathway constraints and exploration logic within the planning space \citep{xie2022retrograph, zhang2025data, sadowski2025trustworthyretrosynthesiseliminatinghallucinations}. Despite these advancements, the architectural decoupling between the predictive engine and the search module remains. This separation confines the planner to fragmented interactions, precluding a continuous logical flow and often leading to a lack of global foresight.

\textbf{Reasoning-enhanced retrosynthetic planning.} The rise of LLMs has shifted the focus toward reasoning-enhanced frameworks that internalize chemical knowledge for strategic planning. Early applications primarily utilized LLMs as sequence-to-sequence engines for single-step prediction \citep{Yang2025BatGPTChemAF, deng2025rsgpt, hassen2025atomanchoredllmsspeakchemistry}. To tackle the complexities of multi-step tasks, subsequent research utilized textual in-context learning to internalize multi-step synthetic dependencies \citep{kang2025retrointext}, or developed agentic workflows where LLMs iteratively orchestrate external tools \citep{liu2025retror, sathyanarayana2025deepretroretrosyntheticpathwaydiscovery, Bran2025ChemicalRI}. Further efforts have sought to enhance the strategic reasoning depth of LLMs, such as utilizing RL to refine the reasoning logic of single-step predictors \citep{zhang2025reasoningdrivenretrosynthesispredictionlarge}, employing evolutionary algorithms for iterative path-level correction \citep{wang2025llmaugmented}, or integrating multimodal graph encoders to provide heuristic guidance for discrete search algorithms \citep{liu2025multimodal}. Despite these developments, most existing systems remain architecturally decoupled, relying on external search or corrective modules to ensure path-level viability \citep{song2025aotefficientsynthesisplanning, sadowski2025trustworthyretrosynthesiseliminatinghallucinations}. ReTriP departs from these decoupled architectures by unifying global planning logic into a standalone CoT process, enabling the model to directly derive rigorous synthetic routes through intrinsic chemical foresight.

\section{Problem formulation}

The objective of retrosynthetic planning is to identify a feasible synthetic pathway that connects a target molecule $T$ to a set of accessible starting materials. Following the convention established by \citet{bradshaw2020barking} and \citet{schwaller2020predicting}, we formalize a synthetic route as a Directed Acyclic Graph (DAG) $\mathcal{G} = (\mathcal{V}, \mathcal{E})$, where the set of vertices $\mathcal{V}$ represents chemical molecules, and the directed edges $\mathcal{E}$ denote reaction transformations. To ensure chemical and logical rigor, $\mathcal{G}$ must satisfy the following properties:

\begin{itemize}
    \item \textbf{Target convergence:} The target molecule $T \in \mathcal{V}$ is the unique sink of $\mathcal{G}$, characterized by an out-degree $d_{out}(T) = 0$. It serves both as the logical origin of the retrosynthetic analysis and the final product of the forward synthesis.
    \item \textbf{Starting material grounding:} Defining the set of leaf nodes as $\mathcal{R} = \{v \in \mathcal{V} \mid d_{in}(v) = 0\}$, we require $\mathcal{R} \subseteq \mathcal{S}$, where $\mathcal{S}$ is a vast predefined library of commercially available materials \citep{sterling2015zinc, genheden2020aizynthfinder}.
    \item \textbf{Stepwise fidelity:} Every non-leaf node $v \in \mathcal{V} \setminus \mathcal{R}$ must correspond to a validated reaction, whereby $v$ is derivable from its immediate children $\mathcal{C}_v = \{u \mid (u, v) \in \mathcal{E}\}$.
\end{itemize}

Ultimately, the core task of retrosynthetic planning is to learn a mapping $T \to \mathcal{G}$ that generates a valid synthetic graph for a given target. Although chemical validity is a strict prerequisite, routes that maximize synthetic efficiency are highly preferred in practice \citep{corey1991logic}. Algorithmically, this efficiency is typically operationalized by bounding the maximum path depth $D_{\mathcal{G}} = \max_{r \in \mathcal{R}} \text{dist}(r, T)$, which acts as a practical proxy for cumulative yield \citep{chen2020retro, liu2023fusionretro}.

\begin{figure}[t]
\begin{center}
\includegraphics[width=1.0\linewidth]{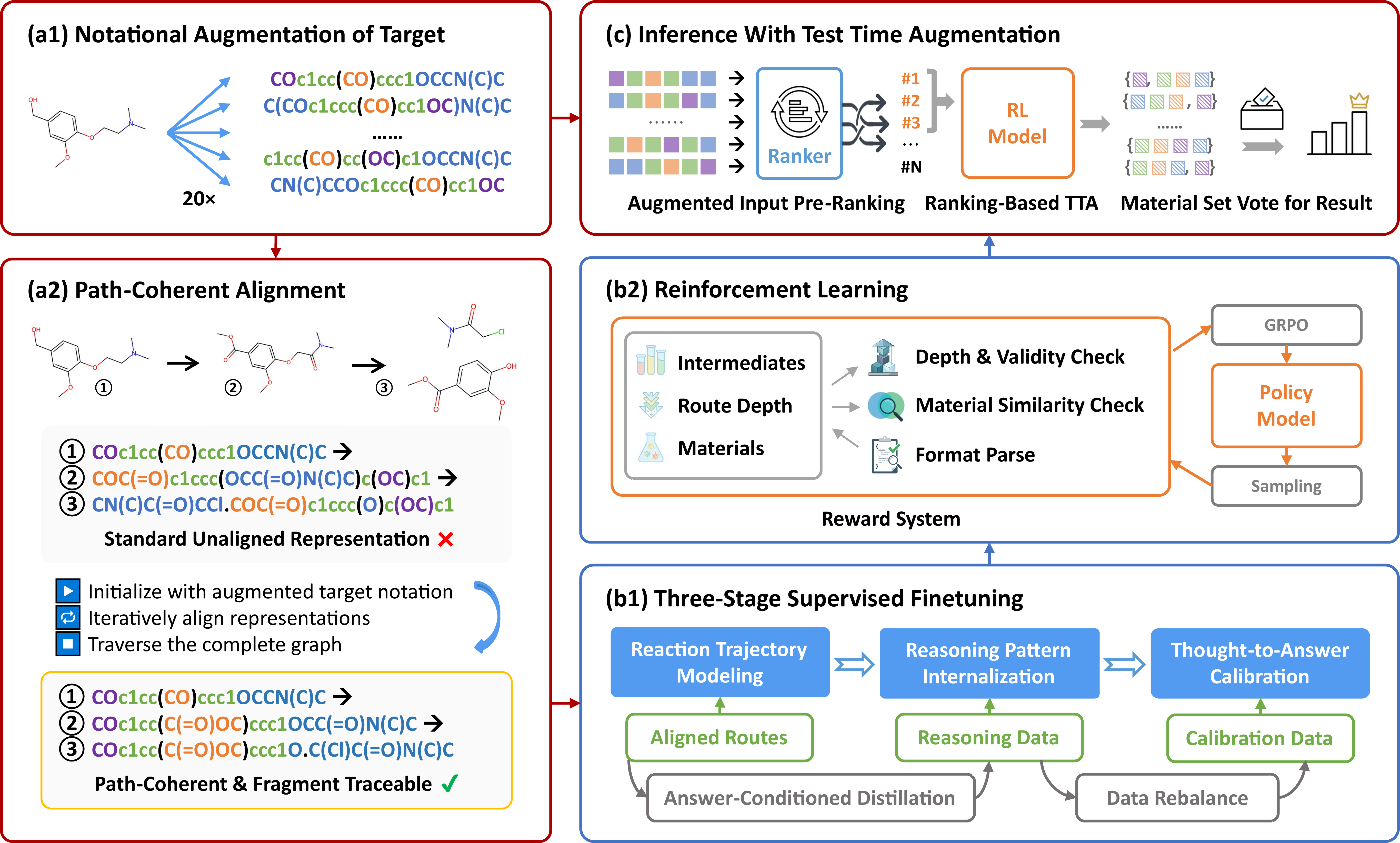}
\caption{\textbf{The ReTriP framework.} (a) Path-coherent data construction: (a1) The target molecule undergoes notational augmentation. (a2) Iterative alignment ensures fragment traceability. (b) Progressive training curriculum: (b1) A three-stage SFT process transitions from trajectory modeling to reasoning distillation and loss-rebalancing calibration. (b2) RLVR aligns stepwise logic with precursor availability and synthetic efficiency. (c) Inference and scaling: A ranking-guided TTA selects optimal input notations, followed by a consensus-based voting mechanism.}
\label{fig:main_pipeline}
\end{center}
\end{figure}

\section{Methodology}

\subsection{Overview of ReTriP}

ReTriP reformulates retrosynthetic planning as an end-to-end sequence generation task (Figure~\ref{fig:main_pipeline}). This framework unifies local reaction logic with global planning objectives by integrating path-coherent molecular representations and a goal-oriented optimization curriculum, directly mapping the target molecule $T$ to synthetic graph $\mathcal{G}$ via LLM CoT reasoning.

The first component of ReTriP mitigates the notational discontinuity of conventional SMILES sequences in multi-step contexts, which obscures underlying topological transformations and introduces syntactic noise \citep{schwaller2021extraction, zhong2022root}. To preserve structural traceability, ReTriP implements path-coherent alignment by anchoring reactant notations to their corresponding fragments in preceding products (Figure~\ref{fig:main_pipeline}a). Additionally, as the initial representation of the target molecule defines the logical frame for all subsequent reasoning \citep{schwaller2019molecular, ordereffect, zhang2025reasoningdrivenretrosynthesispredictionlarge}, the framework utilizes notational pre-ranking (Figure~\ref{fig:main_pipeline}c) to identify the most informative starting point \citep{do2025automatic}. By stabilizing both the initial representational context and the structural correspondence across the route, ReTriP allows the model to focus on reaction logic rather than representation-level fluctuations.

The second component of ReTriP aligns stepwise generation with global synthetic utility through a progressive optimization curriculum (Figure~\ref{fig:main_pipeline}b). Simple imitation learning restricts direct generation to historical biases, failing to align with terminal precursor grounding or minimal route depth \citep{shee2025directmultistep, xuanvu2025tempretemplategenerationsingle}. To address this, ReTriP first employs three-stage supervised fine-tuning (SFT) to internalize inter-step reaction dependencies and the structured reasoning format necessary for long-horizon planning \citep{ouyang2022training, wang2025chemrlearningreasonchemist}. Subsequently, RLVR is introduced to shift the optimization objective toward practical criteria \citep{shao2024deepseekmathpushinglimitsmathematical, liu2025retror, wen2025reinforcementlearningverifiablerewards}. By training against rewards derived from precursor availability and route parsimony, the model learns to prioritize practically preferred routes. This specialized curriculum allows ReTriP to internalize the global dependencies and strategic foresight typically delegated to external search heuristics.

Ultimately, ReTriP operates as a synergistic pipeline where path-coherent representations enable the internalization of global synthetic objectives, allowing for a continuous and chemically grounded logical flow. During inference, this internalized reasoning is further scaled using a ranking-guided Test-Time Augmentation (TTA) with a consensus-based voting scheme to ensure robust path discovery \citep{wang2023selfconsistency} (Figure~\ref{fig:main_pipeline}c).

\subsection{Root-aligned multi-step representation}

To enable end-to-end generation, we represent the synthetic route as a linear sequence of reaction SMILES \citep{xuanvu2025tempretemplategenerationsingle, granqvist2026retrosynformer}. Specifically, we first transform the retrosynthetic DAG into a hierarchical tree $\mathcal{T}$, rooted at the target $T$, by decoupling any convergent paths. The tree $\mathcal{T}$ is subsequently linearized via a depth-first traversal.

Building on this linearized structure, we establish topological correspondence across consecutive reactions. To advance beyond the single-step alignment provided by R-SMILES \citep{zhong2022root}, we introduce a global topological synchronization strategy that propagates structural alignment across the entire multi-step pathway. During transitions, intermediate molecules retain their prior root-atom designations. New precursors are recursively aligned by selecting root atoms mapped to the lowest-index positions in the product's SMILES. This inheritance mechanism transforms the sequence into a coherent structural map, enabling the model to trace fragment evolution without arbitrary atom re-indexing.

The selection of the initial root atom for target $T$ serves as a global anchor for route notation. We leverage this by applying a 20-fold randomized augmentation \citep{smilesaug} during training, exposing the model to identical chemical transformations through diverse notational perspectives \citep{schwaller2019molecular, zhong2022root}. By decoupling fundamental chemical logic from representational biases, we encourage the model to internalize topological invariants rather than memorizing specific string patterns. Such notational robustness is critical for maintaining reasoning integrity throughout long-horizon planning.

\subsection{Progressive internalization of planning logic}

To ground the planning process in a robust understanding of chemical space, ReTriP is initialized from a pre-trained foundational molecular LLM. While such models \citep{zhang2024chemllm, zhao2025developing, zuo2025biomedgpt} possess extensive knowledge of basic reaction rules, they often lack the path-level perspective required for long-horizon retrosynthetic coordination \citep{valmeekam2023planning, bran2025chemicalreasoningllmsunlocks}. 
We address this limitation through a three-stage SFT curriculum (Figure~\ref{fig:main_pipeline}b1) that systematically transforms static molecular knowledge into a dynamic, strategy-driven planning process.

\paragraph{Stage 1: Reaction trajectory modeling.} 
The first stage establishes path-level structural priors by training the model on the proposed path-coherent route representation. This process internalizes the long-range, stepwise dependencies inherent in multi-step routes. By capturing the statistical distribution of valid trajectories, the model develops a foundational understanding of synthetic viability before explicit reasoning patterns are introduced.

\paragraph{Stage 2: Reasoning pattern internalization.} 
In this stage, we incorporate explicit CoT reasoning via answer-conditioned distillation \citep{hsieh2023distilling} from a general-purpose LLM. By conditioning the distillation prompt on a ground-truth route, we enable the teacher model to articulate the specific, step-by-step chemical logic \citep{corey1991logic} underlying each pathway. These distilled CoT trajectories explicitly rationalize strategic disconnections and anticipate reactivity constraints, such as functional group compatibility or steric hindrance. By internalizing these reasoning patterns, ReTriP develops deep chemical foresight, ensuring its generative process is governed by rigorous rationale rather than superficial pattern matching.

\paragraph{Stage 3: Thought-to-answer calibration.} 
The final stage addresses the token-level statistical disparity between natural language reasoning and molecular notation. Since linguistic tokens typically exhibit higher entropy than the highly constrained SMILES syntax, a standard joint optimization can lead to a gradient imbalance where the model over-prioritizes the stylistic patterns of reasoning at the expense of notational precision \citep{hao2025funreasonenhancinglargelanguage, fang2025thinkless}. To achieve supervision rebalance, we adopt a weighted loss function:
\begin{equation}
\mathcal{L}_{total} = \alpha \cdot \mathcal{L}_{thought} + (1 - \alpha) \cdot \mathcal{L}_{answer}
\end{equation}
Specifically, $\mathcal{L}_{thought}$ and $\mathcal{L}_{answer}$ denote the average per-token cross-entropy loss for the reasoning and answer segments, respectively, while $\alpha$ serves as a balancing hyperparameter. By balancing these gradients, ReTriP maintains high-fidelity molecular generation while preserving the depth of its latent reasoning trajectories.

\subsection{Strategic alignment with global planning goals}

Building on the foundational logic established through SFT, we utilize RLVR to transition the model toward autonomous strategic exploration (Figure~\ref{fig:main_pipeline}b2). Retrosynthesis is inherently a one-to-many problem where multiple valid pathways may exist for a single target. To navigate this high-dimensional space, we employ a hierarchical reward function \citep{touvron2023llama2openfoundation, lai2024alarm, Guo_2025} $R$ designed to offer fine-grained feedback based on objective measures of synthetic success. 

We define the complete identification of a precursor set $\mathcal{R}$ as the pivotal goal. This grounding constraint acts as a robust indicator of route validity \citep{liu2023fusionretro, kang2025retrointext}.

To ensure that format adherence and precursor grounding serve as the primary drivers of the optimization before the model accounts for route efficiency, $R$ is implemented as a conditional gated sequence. Let $\mathcal{G}$ be the generated plan. The total reward is formulated as:
\begin{equation}
R = 
\begin{cases} 
0 & \text{if } \mathcal{G} \text{ is unparsable} \\
R_{\text{fmt}} + \Phi(\mathcal{G}) & \text{if } \mathcal{G} \text{ is parsable}
\end{cases}
\end{equation}
where $R_{\text{fmt}} = 0.5$ denotes a structural adherence bonus, incentivizing the correct use of reasoning delimiters and the syntactic layout of the multi-step reaction SMILES. The potential function $\Phi(\mathcal{G})$ provides a holistic assessment of planning quality, primarily driven by the matching of the proposed precursor set $\mathcal{R}$ against the collection of all feasible ground-truth sets $\smash{\{\mathcal{R}^*_j\}}$. Specifically, we define the maximum Jaccard similarity as $\smash{J_{\max} = \max_j J(\mathcal{R}, \mathcal{R}^*_j)}$ and formulate $\smash{\Phi(\mathcal{G})}$ to differentiate between exact goal attainment and partial progress:
\begin{equation}
\Phi(\mathcal{G}) = 
\begin{cases} 
\lambda_{\text{acc}} - \Psi(\mathcal{G}) & \text{if } J_{\max} = 1 \\
\lambda_{\text{sim}} \cdot J_{\max} & \text{otherwise}
\end{cases}
\end{equation}
Here, $\lambda_{\text{acc}} = 1.5$ is the reward for matching any feasible retrosynthetic target, while $\lambda_{\text{sim}} = 0.5$ provides a gradient for partial matching to prevent reward sparsity.

For successful routes where $\mathcal{R}$ exactly matches any ground-truth set $\smash{\mathcal{R}^*_j}$, we introduce a refinement penalty $\Psi(\mathcal{G})$ to incentivize chemical rigor and route parsimony. This penalty is applied conditionally to avoid hindering the initial exploration of starting materials:
\begin{equation}
\Psi(\mathcal{G}) = \eta_{v} \min(c_{\text{inv}}, 4) + \eta_{d} ~\text{clip}(D_{\mathcal{G}} - D_{\mathcal{G}}^*, 0, 3)
\end{equation}
where $c_{\text{inv}}$ counts reactions with invalid SMILES and $\text{clip}(\cdot, 0, 3)$ bounds the excess depth relative to the reference depth $\smash{D_{\mathcal{G}}^*}$. We set $\smash{\eta_{v} = 0.1}$ and $\smash{\eta_{d} = 0.2}$ to satisfy $\smash{\lambda_{acc} - \max(\Psi) \ge \lambda_{sim}}$, ensuring that exact precursor recovery consistently carries a greater incentive than partial matches, regardless of route complexity.

\subsection{Notational prioritization and inference scaling}

To mitigate the notational variance inherent in molecular notations and expand the exploration during generation, we implement a two-stage inference strategy (Figure~\ref{fig:main_pipeline}c). This approach optimizes the initial representational context and utilizes inference-time scaling to ensure the model's trajectory is grounded in the most informative structural perspective.

\textbf{Notational pre-ranking.} The first stage employs a lightweight scoring module to identify the optimal starting representation for target $T$ by evaluating equivalent notational variants. This scorer consists of a MLP-based ranking head atop a pre-trained molecular representation model \citep{edwards-etal-2022-translation, chemberta, ross2022large}, fine-tuned on the empirical planning outcomes of the LLM. By assessing the planning feasibility of various root-atom anchorings, the scorer selects the top-ranked SMILES as LLM input. This mechanism suppresses notational noise that could otherwise precipitate suboptimal routes.

\textbf{Consensus-based test-time augmentation.} To leverage representational diversity for broader path coverage, we implement a multi-view scaling strategy via TTA \citep{snell2025scaling}. Rather than relying on a single deterministic trajectory, ReTriP performs parallel, fully independent inference across the top-$k$ ranked SMILES identified in the first stage. This enables the model to exploit distinct reasoning branches induced by varied initial notations. To select the most robust trajectory from these outputs, we extract the terminal precursor set $\mathcal{R}$ from each plan and apply a frequency-based majority vote \citep{wang2023selfconsistency, snell2025scaling}. We identify the final solutions by directly ranking these independent candidates according to their precursor consensus scores. This strategy significantly enhances the discovery of valid routes for long-horizon planning.

\section{Experiments}

\subsection{Experimental setup}

\textbf{Dataset.} We evaluate ReTriP on RetroBench \citep{liu2023fusionretro}, a rigorous benchmark for multi-step retrosynthetic planning derived from the USPTO-full reaction network \citep{li2022prediction, chen2020retro}. The dataset consists of 46,458 training, 5,803 validation, and 5,838 testing target molecules, with synthetic routes spanning a depth of 2 to 13 steps. We adopt identical data partitions for both the generative planner and the notational pre-ranking module. Through 20-fold root-aligned augmentation, we expand the training set to approximately 929k trajectories.

\textbf{Implementation.} We initialize our generative planner with the pre-trained weights of BioMedGPT-Mol \citep{zuo2025biomedgpt}, which is adapted from Qwen3-8B \citep{qwen3}. The SFT curriculum is implemented using LoRA with a rank of 16, where each training stage is conducted for 1.5 epochs. We fine-tune the model on high-quality reasoning data distilled from DeepSeek-V3.1 \citep{deepseekai2024deepseekv3technicalreport}. In the third SFT stage, we set the rebalancing coefficient $\alpha$ to 0.1. For the RL phase, we employ a GRPO \citep{shao2024deepseekmathpushinglimitsmathematical} variant utilizing the DAPO loss \citep{yu2025dapo}, performing full-parameter fine-tuning. The notational pre-ranking module is based on the MolFormer-XL \citep{ross2022large} architecture with 44M parameters. Empirically, training the scorer to full convergence maximizes its inference-time robustness. Detailed implementation configurations are provided in the Appendix.

\textbf{Baselines.} We evaluate ReTriP against a comprehensive suite of baselines, encompassing template-based models such as Retrosim \citep{coley2017computer} and Neuralsym \citep{segler2017neural}, semi-template approaches like GraphRetro \citep{somnath2021learning} and G2Gs\citep{shi2020graph}, and template-free methods including Megan \citep{sacha2021molecule} and FusionRetro \citep{liu2023fusionretro}. Our comparison also includes the current SOTA LLM-enhanced multimodal framework RetroInText \citep{kang2025retrointext} as well as reranked variants like FusionRetro combined with CREBM \citep{crebm}. Since these methods generally rely on external search algorithms like Retro* or Retro*-0 \citep{chen2020retro} for multi-step planning, we report their optimal search-integrated results. In contrast, ReTriP operates as a standalone generator, with its performance reported using 16-fold TTA.

\textbf{Evaluation metrics.} Following the established protocol on RetroBench, we employ top-$k$ exact match accuracy as our primary performance metric. A synthetic route is deemed successful only if the proposed set of precursors matches a verified ground-truth set without exceeding the reference path depth.

\begin{table}[t]
\begin{center}
\begin{small}
\setlength{\tabcolsep}{4pt}
\begin{tabular}{llccccc}
\toprule
\textbf{Generative Model} & \textbf{Search Module} & \textbf{Top-1} & \textbf{Top-2} & \textbf{Top-3} & \textbf{Top-4} & \textbf{Top-5} \\ \midrule
\multicolumn{1}{c}{Template-based} & & & & & & \\
\midrule
Retrosim \citep{coley2017computer} & Retro* & 35.1 & 40.5 & 42.9 & 44.0 & 44.6 \\
Neuralsym \citep{segler2017neural} & Retro*-0 & 42.0 & 49.3 & 52.0 & 53.6 & 54.3 \\
GLN \citep{gln} & Retro* & 39.6 & 48.9 & 52.7 & 54.6 & 55.7 \\
\midrule
\multicolumn{1}{c}{Semi-template-based} & & & & & & \\
\midrule
G2Gs \citep{shi2020graph} & Retro* & 5.4 & 8.3 & 9.9 & 10.9 & 11.7 \\
GraphRetro \citep{somnath2021learning} & Retro* & 15.3 & 19.5 & 21.0 & 21.9 & 22.4 \\
GraphRetro+CREBM\citep{crebm} & Retro*-0 & 16.3 & 20.2 & 21.6 & 22.3 & 22.7 \\
\midrule
\multicolumn{1}{c}{Template-free} & & & & & & \\
\midrule
Transformer \citep{karpov2019transformer} & Retro* & 31.3 & 40.4 & 44.7 & 47.2 & 48.9 \\
Transformer+CREBM \citep{crebm} & Retro* & 35.0 & 43.4 & 46.7 & 48.5 & 49.7 \\
Megan \citep{sacha2021molecule} & Greedy DFS & 32.9 & -- & -- & -- & -- \\
FusionRetro \citep{liu2023fusionretro} & Retro* & 37.5 & 45.0 & 48.2 & 50.0 & 50.9 \\
FusionRetro+CREBM \citep{crebm} & Retro*-0 & 39.6 & 46.7 & 49.5 & 51.0 & 51.7 \\
RetroInText \citep{kang2025retrointext} & Retro*-0 & 42.1 & 49.9 & 53.0 & 54.7 & 55.7 \\
\textbf{ReTriP (Ours)} & \textbf{N/A} & \textbf{45.5} & \textbf{52.1} & \textbf{55.1} & \textbf{56.7} & \textbf{57.6} \\
\bottomrule
\end{tabular}
\end{small}
\caption{\textbf{Exact match accuracy (\%) on RetroBench.} Baseline results are as reported by \citet{liu2023fusionretro}, \citet{crebm}, and \citet{kang2025retrointext}.}
\label{tab:main_results}
\end{center}
\end{table}

\subsection{Performance on RetroBench}

\textbf{Comparison with baselines.} The comparative performance of ReTriP against baselines on RetroBench is summarized in Table~\ref{tab:main_results}. Our model achieves a Top-1 accuracy of 45.5\%, representing a substantial margin of improvement (3.4\%) over the previous multimodal SOTA, RetroInText + Retro*-0. While traditional frameworks rely on external search, ReTriP maintains higher planning accuracy as a standalone planner, validating the effectiveness of internalizing long-range chemical foresight.



\begin{figure}[t]
    \begin{center}
    \begin{minipage}[t]{0.48\textwidth} 
        \centering
        \vspace{0pt} 
        \includegraphics[width=\linewidth]{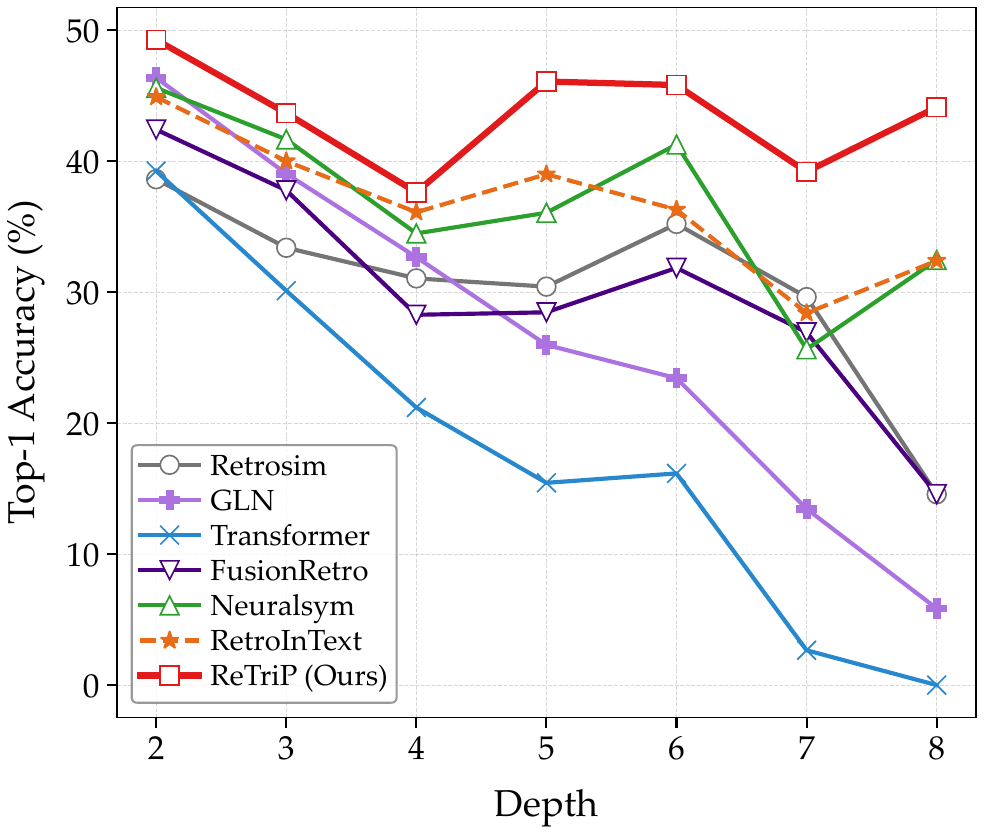}
        \caption{\textbf{Top-1 accuracy across different synthetic route depths.}}
        \label{fig:depth_analysis}
    \end{minipage}
    \hfill 
    \begin{minipage}[t]{0.48\textwidth}
        \centering
        \vspace{0pt} 
        \includegraphics[width=\linewidth]{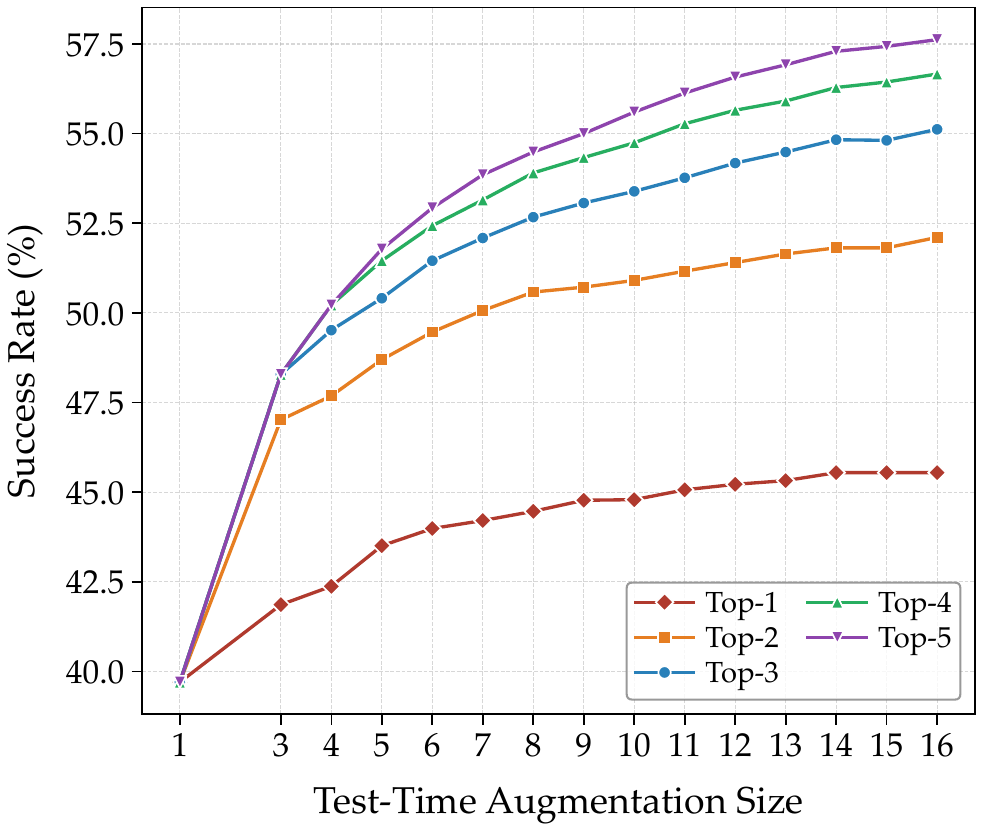}
        \caption{\textbf{Top-k accuracy of the final model across different TTA sizes.}}
        \label{fig:topk_scaling}
    \end{minipage}
    \end{center}
\end{figure}

\textbf{Analysis for the depth of routes.} We evaluate long-range planning robustness by analyzing Top-1 accuracy across varying path depths (Figure~\ref{fig:depth_analysis}), comparing ReTriP against the representative Retro* search algorithm combined with diverse single-step engines. The results demonstrate that while Retro*-based methods suffer from performance degradation as route depth increases, ReTriP remains remarkably robust, maintaining approximately 40\% accuracy even for paths with depth $D_{\mathcal{G}} \ge 5$. Notably, as 75\% of RetroBench test samples are shallow ($D_{\mathcal{G}} \le 3$), standard aggregate metrics potentially understate ReTriP's strategic advantage in complex scenarios. These findings validate that ReTriP navigates retrosynthetic spaces more effectively than traditional search heuristics.

\begin{table}[t]
\begin{center}
\begin{tabular}{cccc|cccc}
\toprule
\multicolumn{4}{c}{\textbf{Components}} & \multicolumn{4}{c}{\textbf{Accuracy (\%) @ TTA Size}} \\
\cmidrule(lr){1-4} \cmidrule(lr){5-8}
\textbf{Path Alignment} & \textbf{CoT} & \textbf{RLVR} & \textbf{Pre-ranking} & \textbf{1} & \textbf{3} & \textbf{4} & \textbf{5} \\
\midrule
& & & & 30.3 & 32.8 & 34.0 & 35.0 \\
$\checkmark$ & & & & 33.5 & 36.1 & 37.6 & 38.5 \\
$\checkmark$ & $\checkmark$ & & & 33.9 & 37.3 & 39.2 & 40.6 \\
$\checkmark$ & $\checkmark$ & $\checkmark$ & & 36.6 & 39.3 & 40.6 & 41.7 \\
$\checkmark$ & $\checkmark$ & $\checkmark$ & $\checkmark$ & \textbf{39.7} & \textbf{41.9} & \textbf{42.4} & \textbf{43.5} \\
\bottomrule
\end{tabular}
\caption{\textbf{Component-wise ablation study.} Without pre-ranking, inference inputs are randomly sampled. Stochastic results report the mean of 10 runs.}
\label{tab:method_ablation}
\end{center}
\end{table}

\subsection{Ablation experiments}

To assess the impact of representational diversity, we evaluate the performance gains achieved by TTA. As shown in Figure~\ref{fig:topk_scaling}, all top-$k$ accuracy metrics improve monotonically with TTA size. Specifically, Top-1 accuracy increases from 39.7\% (size 1) to 45.5\% (size 16), while Top-5 reaches 57.6\%. The sustained performance growth implies that the model has not merely memorized a single optimal path, but has internalized a broad distribution of valid chemical strategies that can be uncovered through representational exploration.

To isolate the performance drivers within ReTriP, we perform a component-wise ablation study, as summarized in Table~\ref{tab:method_ablation}. The pre-trained BioMedGPT-Mol \citep{zuo2025biomedgpt} model yields 0\% accuracy under both zero-shot and one-shot settings, confirming its native inability to perform multi-step planning. To establish a practical starting point, we train a baseline restricted to intra-reaction R-SMILES with randomized inter-step roots. Compared to this baseline, our path-coherent alignment delivers a solid +3.2\% gain (TTA-1) by preserving structural traceability across the entire synthetic route. Building upon this structural foundation, the integration of CoT-SFT contributes an additional 2.1\% boost at TTA-5. This indicates that explicit rationales are crucial for navigating complex decision branches, thereby enhancing the model's strategic diversity. Subsequently, RL alignment provides a significant uplift, elevating TTA-1 accuracy to 36.6\% and validating the transition from localized imitation to global objective alignment. Finally, notational pre-ranking delivers a decisive +3.1\% gain (TTA-1), proving that an optimal representational vantage point is essential for navigating the planning space.

\section{Conclusion}

We present ReTriP, a standalone generative framework that internalizes multi-step retrosynthetic planning within a unified CoT process. By integrating path-coherent representations with strategic objective alignment, we bridge the logical gap of conventional decoupled systems. Our experiments on RetroBench demonstrate that ReTriP achieves SOTA performance and exhibits superior robustness in navigating deep combinatorial spaces where traditional search algorithms falter. Complementing its predictive accuracy, the generated rationales provide interpretable strategies that facilitate manual verification. This shift to intrinsic chemical reasoning establishes a new paradigm for autonomous retrosynthesis. Ultimately, ReTriP provides a scalable foundation for future autonomous chemical agents, promising to accelerate the discovery and synthesis of complex molecular architectures. Future work will explore the framework’s extensibility to broader chemical domains.

\section*{Ethics Statement}

\paragraph{Broader impact and dual-use risks.}
This paper introduces ReTriP, a computational framework trained on standard, publicly available datasets to accelerate legitimate chemical research and drug discovery. While we acknowledge that generative chemical models inherently carry dual-use risks (e.g., identifying synthetic routes for restricted substances), ReTriP is strictly a computational tool. It does not bypass the stringent real-world regulatory and physical barriers required for chemical procurement. We emphasize that this framework is designed to augment, not replace, human chemists. Any practical application of ReTriP must incorporate human-in-the-loop oversight and rigorous safety assessments.

\paragraph{AI usage disclosure.}
During the preparation of this manuscript, the authors utilized generative artificial intelligence tools for two purposes: (1) to assist with language translation and text polishing of the final manuscript; and (2) to generate certain graphical assets, specifically the icons used in Figures~\ref{fig:paradigm_single} and \ref{fig:main_pipeline}. The human authors carefully reviewed, selected, and integrated all AI-generated content, and take full and sole responsibility for the original ideas, data, scientific accuracy, visual representation, and final content of this paper.

\bibliography{main}
\bibliographystyle{colm2026_conference}

\appendix

\section{Native retrosynthetic capabilities of LLMs}

Recent literature has demonstrated that despite their broad chemical knowledge, general LLMs struggle with out-of-the-box retrosynthesis \citep{li2025retroexpertcollaborativereasoninginterpretable, hassen2025atomanchoredllmsspeakchemistry, zhang2025reasoningdrivenretrosynthesispredictionlarge}. Building upon these established findings, we provide an additional quantitative baseline specifically on the RetroBench dataset to properly contextualize the architectural advancements of ReTriP.

\textbf{Experimental setup.} Evaluated models include advanced general LLMs (DeepSeek-V3.2 \citep{deepseekai2025deepseekv32pushingfrontieropen}, GLM-5 \citep{glm5team2026glm5vibecodingagentic}, Kimi-K2.5 \citep{kimiteam2026kimik25visualagentic}, Qwen3.5-397B-A17B \citep{qwen3.5}) and specialized chemistry models (ChemDFM-R-14B \citep{zhao2025chemdfm}, ChemLLM-20B-Chat-DPO \citep{zhang2024chemllm}, and our base model BioMedGPT-Mol \citep{zuo2025biomedgpt}). 

To ensure a fair comparison with ReTriP's unaugmented, single-pass generation (TTA size of 1), all baselines were restricted to a single inference pass. We evaluated the models across several configurations, specifically testing one-shot and 3-shot in-context learning settings using exemplars of varying retrosynthetic complexity. To mitigate the verbosity of instruction-tuned models, task-specific system prompts enforced a JSON output schema. We experimented with sampling temperatures of 0.1, 0.2, and 0.5, reporting the best-observed performance for each model. Furthermore, all models were evaluated using their explicit reasoning or "thinking" modes where supported.

\textbf{Results and discussion.} As shown in Table~\ref{tab:llm_baselines}, despite massive parameter scales, all baseline models exhibit a significant performance deficit, peaking at 1.9\% single-pass accuracy. While this is partly due to unfamiliarity with the specific data distribution and starting material definitions in RetroBench, the results underscore that general-purpose capabilities alone are insufficient for complex synthetic coordination. In contrast, ReTriP achieves 39.7\% accuracy under identical constraints, demonstrating that our path-coherent representations and RLVR alignment effectively internalize the domain-specific logic and constraints necessary for viable planning.

\begin{table}[htbp]
\begin{center}
\begin{small}
\begin{tabular}{llcc}
\toprule
\textbf{Model Category} & \textbf{Model} & \textbf{Parameters} & \textbf{Single-pass Acc (\%)} \\ 
\midrule
\multirow{4}{*}{General LLMs} 
& GLM-5 & 744B & $1.8$ \\
& DeepSeek-V3.2 & 671B & $1.9$ \\
& Qwen3.5-397B-A17B & 397B & $1.6$ \\
& Kimi-K2.5 & 1.04T & $1.0$ \\
\midrule
\multirow{3}{*}{Chemistry LLMs} 
& ChemLLM-20B-Chat-DPO & 20B & $0.0$ \\
& ChemDFM-R-14B & 14B & $0.5$ \\
& BioMedGPT-Mol (ReTriP Base) & 8B & $0.0$ \\
\midrule
\textbf{Our Framework} & \textbf{ReTriP (TTA Size = 1)} & \textbf{8B} & \textbf{39.7} \\
\bottomrule
\end{tabular}
\end{small}
\caption{\textbf{Out-of-the-box performance of foundation models on RetroBench.}}
\label{tab:llm_baselines}
\end{center}
\end{table}

\paragraph{Rationale for selecting BioMedGPT-Mol.} Although BioMedGPT-Mol lacks native multi-step planning capabilities, we empirically found that its molecular-level pre-training provides a superior foundation for retrosynthetic alignment. Initializing Stage 1 SFT from a vanilla Qwen3-8B \citep{qwen3} instead of BioMedGPT-Mol resulted in a decrease in single-pass accuracy from 33.5\% to 30.3\%. This performance gap suggests that the internalized knowledge of SMILES syntax and functional group chemistry provides a critical structural prior for acquiring complex planning logic.

\section{Rationale for benchmark and evaluation metric selection}

Conventionally, the evaluation of multi-step retrosynthetic algorithms relies on the search success rate \citep{chen2020retro, genheden2022paroutes, yuan2024active, liu2025retror, wang2025llmaugmented}. However, this metric exhibits a critical flaw. Integrating single-step models that achieve 60 to 80 percent top-5 accuracy with search algorithms such as Retro* \citep{chen2020retro} paradoxically inflates multi-step success rates to between 85 and 94 percent \citep{liu2023fusionretro}. This anomaly arises because the traditional metric inherently credits any pathway terminating in available materials. It fails to verify the actual chemical viability of intermediate reactions, thereby accumulating false positives.

To address this limitation, we adopt the RetroBench benchmark \citep{liu2023fusionretro} and its stringent set-wise exact match metric. Under this protocol, a route is deemed successful only if the completely predicted set of starting materials perfectly matches a ground-truth reference set. Furthermore, unconstrained expansions are prevented by strictly bounding the maximum search depth to the length of the reference route. This framework resolves the structural deficiencies of conventional metrics, ensuring that our evaluation accurately reflects practical synthetic viability.

\section{Inference efficiency and computational cost}

We benchmark the inference efficiency of ReTriP using a vLLM server deployed on a single NVIDIA A800 GPU. In a throughput-optimized scenario with 1024 concurrent API requests, evaluating the entire RetroBench test set (TTA-1) takes 1118.3 seconds, yielding an amortized cost of 191.6 ms per target. Alternatively, to prioritize individual request latency, accessing the API with 16 concurrent threads results in an absolute latency of approximately 11.7 seconds. Since our TTA branches are fully independent, a complete 16-fold TTA evaluation for a single target can be executed in parallel within this 11.7-second window. Given the minute-to-hour computational budgets typical of conventional synthesis planning, such latency is practically negligible.

Ultimately, ReTriP introduces a fundamentally more capable planning paradigm rather than merely optimizing for speed. There is a profound distinction in how ReTriP and traditional decoupled methods scale with computation. Given infinite search time, conventional heuristics merely increase the likelihood of discovering \emph{any} valid pathway, lacking intrinsic mechanisms to gauge holistic route quality. Conversely, our consensus-based TTA explicitly trades parallel compute for genuine accuracy improvements. By aggregating independent reasoning trajectories, ReTriP reliably isolates the highest-quality route rather than merely accumulating valid candidates, ensuring that scaled computation directly translates to verifiable chemical utility.

\section{Qualitative case study}

\begin{figure}[htb]
\begin{center}
\includegraphics[width=1.0\linewidth]{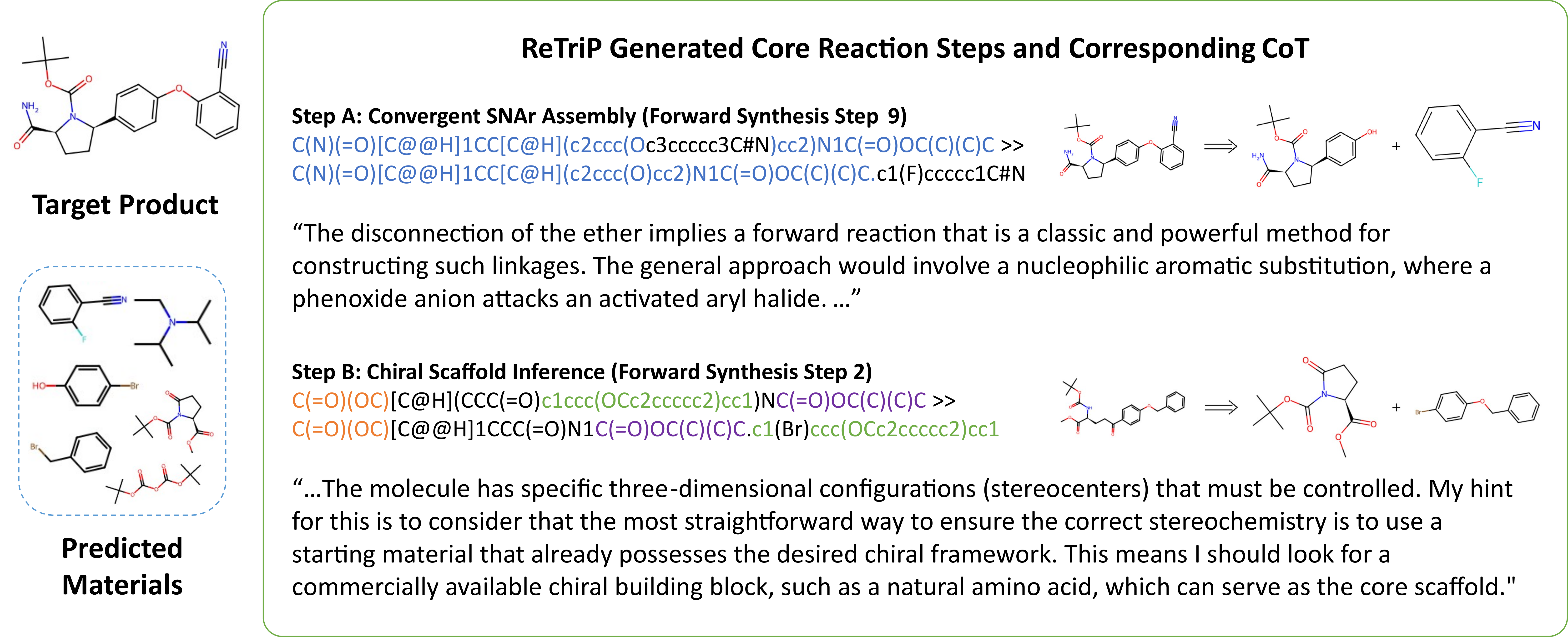} 
\caption{\textbf{Qualitative case study.} A generated 9-step retrosynthetic plan, featuring integrated CoT reasoning for critical steps.}
\label{fig:case_study}
\end{center}
\end{figure}

To qualitatively assess ReTriP's CoT reasoning, we examine a successful 9-step retrosynthetic prediction for a complex target (Figure~\ref{fig:case_study}). Demonstrating both strategic foresight and tactical depth, the framework grounds each transformation in rigorous chemical logic. For instance, in Step A, the model prioritizes an aryl ether disconnection to facilitate a convergent SNAr strategy, while Step B elegantly resolves a stereochemical core by tracing it back to an accessible chiral pool precursor. Crucially, the generated reaction SMILES strictly adhere to our path-coherent notation. This structural fidelity confirms that ReTriP's optimization for synthetic utility is intrinsically bound to the underlying representation, ensuring the generative process remains deeply anchored in chemically traceable logic.

\section{Implementation details}
The development of ReTriP involves two training stages: reasoning-driven supervised fine-tuning (SFT) and goal-oriented reinforcement learning (RL). All experiments are conducted on a cluster of four NVIDIA Tesla A800 GPUs.

\subsection{Supervised fine-tuning stage}

The SFT curriculum comprises three substages: reaction trajectory modeling, reasoning pattern internalization, and thought-to-answer calibration. The model is fine-tuned for $1.5$ epochs in each substage. The corresponding hyperparameters are reported in Table~\ref{tab:sft_param}. In the third substage, the rebalancing coefficient $\alpha$ is set to $0.1$ to prioritize accuracy in the predicted answer segment.

\begin{table}[h]
\begin{center}
\begin{small}
\begin{tabular}{ccccc}
\toprule
\textbf{Hyperparameter} & Epochs & Batch Size & Learning Rate & LoRA Rank \\
\textbf{Values}     & 1.5    &      256      &        1e-4       & 16 \\
\bottomrule
\end{tabular}
\caption{\textbf{Hyperparameters for each supervised fine-tuning stage.}}
\label{tab:sft_param}
\end{small}
\end{center}
\end{table}

\subsection{Reinforcement learning stage}

During the RL phase, we employ Group Relative Policy Optimization (GRPO) \citep{shao2024deepseekmathpushinglimitsmathematical} incorporating the DAPO \citep{yu2025dapo} loss. For each prompt, we sample eight responses to compute group-relative advantages, with a fixed KL regularization coefficient $\smash{\beta}$ used to constrain policy updates. The policy is optimized using a learning rate of $\smash{1 \times 10^{-6}}$ and a batch size of $512$. Comprehensive hyperparameter details are provided in Table~\ref{tab:rl_param}.

\begin{table}[h]
\begin{center}
\begin{small}
\begin{tabular}{ccccccc}
\toprule
\textbf{Hyperparameter} & Group Size & KL $\beta$ & Batch Size & Learning Rate & Temperature & Top-$p$ \\
\textbf{Values}     &   8   &      0.04      &   512   &    1e-6    & 1.0 & 0.95\\
\bottomrule
\end{tabular}
\caption{\textbf{Hyperparameters for the reinforcement learning stage.}}
\label{tab:rl_param}
\end{small}
\end{center}
\end{table}

\subsection{Implementation of notational pre-ranking module}

To mitigate the sensitivity of the LLM to the initial molecular representation, we implement a lightweight notational pre-ranking module. This module identifies the most informative root-aligned SMILES string from a set of equivalent notational variants for a given target molecule $T$.

\paragraph{Model architecture.} 
The ranker is built upon the MoLFormer-XL architecture \citep{ross2022large}, a Transformer-based model pre-trained on a massive chemical corpus. We employ the pre-trained encoder to extract high-dimensional molecular embeddings, specifically utilizing the pooled output from the first token position as the global representation. A task-specific scoring head is appended to the encoder, consisting of a two-layer multi-layer perceptron (MLP) with a hidden dimension of 256, ReLU activation, and a dropout rate of 0.1. The final output is a scalar score $s \in \mathbb{R}$ representing the predicted planning viability of the input notation.

\paragraph{Training objective.} 
The module is trained using a pairwise ranking paradigm. We construct a dataset of SMILES pairs $(x_{pos}, x_{neg})$, where $x_{pos}$ denotes a root-aligned notation that successfully leads to a valid synthetic route in the LLM's empirical planning, while $x_{neg}$ denotes a variant that results in a planning failure. We optimize the model using the Margin Ranking Loss:
\begin{equation}
    \mathcal{L}_{rank} = \max(0, -y \cdot (s_{pos} - s_{neg}) + \text{margin})
\end{equation}
where $y=1$ is the target, $s_{pos}$ and $s_{neg}$ are the predicted scores for the positive and negative notations, respectively, and the margin is set to 1.0. This objective encourages the model to assign higher scores to notations that facilitate successful reasoning trajectories.

\paragraph{Optimization details.} 
The ranking module is fine-tuned using the AdamW optimizer \citep{loshchilov2018decoupled} with a weight decay of 0.01. We utilize a learning rate of $\smash{2 \times 10^{-5}}$ and a global batch size of 128. To ensure training stability, we implement a linear learning rate warmup for the first half-epoch, followed by cosine annealing. Gradient clipping with a maximum norm of 1.0 is applied to prevent divergent updates.

\section{Additional training analysis}
\label{app:training_analysis}

\begin{figure}[htb]
    \centering
    \includegraphics[width=0.55\textwidth]{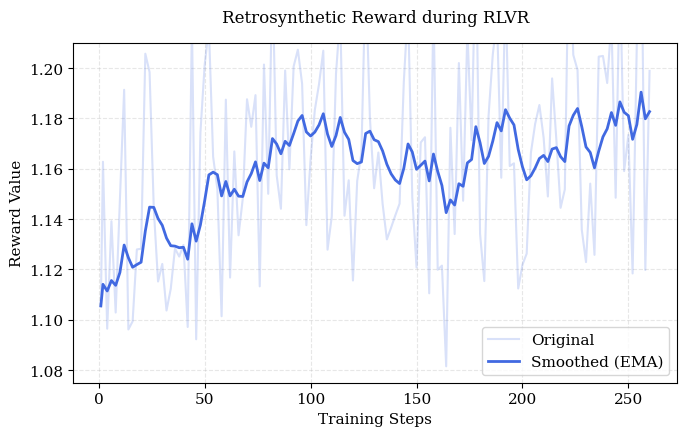}
    \caption{\textbf{Training reward curve of ReTriP during the RLVR phase.} The evolution of the mean retrosynthetic reward demonstrates rapid convergence and effective objective alignment.}
    \label{fig:rl_reward}
\end{figure}

As shown in Figure~\ref{fig:rl_reward}, the model quickly learns to optimize for terminal synthetic success during reinforcement learning, internalizing the global planning constraints.

\section{Path-coherent representation details}
\label{app:path_coherent_details}

In this section, we provide the implementation details of our path-coherent alignment algorithm and demonstrate its superiority over standard canonical representations comprehensively.

\begin{algorithm}[htb]
\caption{Path-coherent sequence alignment.}\label{alg:alignment}
\KwIn{Target molecule $T$, initial root index $r_0 \in \mathcal{V}(T)$, and ordered transformations $\mathcal{X} = \{x_1, \dots, x_n\}$, where each $x_k = (p_k, \mathcal{C}_k, \Phi_k)$ consists of a product $p_k$, precursors $\mathcal{C}_k$, and an atom mapping $\Phi_k : \mathcal{V}(\mathcal{C}_k) \to \mathcal{V}(p_k) \cup \{ \perp \}$.}
\KwOut{Aligned reaction SMILES sequence $\mathcal{A}$.}

\textbf{Initialize:} $RootMap \leftarrow \{T: r_0\}$, $\mathcal{A} \leftarrow \emptyset$\;
\ForEach{transformation $(p, \mathcal{C}, \Phi) \in \mathcal{X}$}{
    $r_p \leftarrow RootMap[p]$\;
    $S_p \leftarrow \text{RootedSMILES}(p, r_p)$\;
    $L_{temp} \leftarrow \emptyset$\;
    \ForEach{precursor $c_i \in \mathcal{C}$}{
        $\mathcal{V}_{mapped} \leftarrow \{ v \in \mathcal{V}(c_i) \mid \Phi(v) \neq \perp \}$\;
        \eIf{$\mathcal{V}_{mapped} \neq \emptyset$}{
            $r_{c_i} \leftarrow \text{argmin}_{v \in \mathcal{V}_{mapped}} \{ \text{Pos}(\Phi(v), S_p) \}$\;
            $p_{c_i} \leftarrow \min_{v \in \mathcal{V}_{mapped}} \{ \text{Pos}(\Phi(v), S_p) \}$\;
        }{
            $r_{c_i} \leftarrow \text{DefaultRoot}(c_i)$, $p_{c_i} \leftarrow \infty$\;
        }
        $RootMap[c_i] \leftarrow r_{c_i}$\;
        $S_{c_i} \leftarrow \text{RootedSMILES}(c_i, r_{c_i})$\;
        $L_{temp} \leftarrow L_{temp} \cup \{(S_{c_i}, p_{c_i})\}$\;
    }
    $\mathcal{C}_{aligned} \leftarrow \text{Sort } \{S_{c_i} \mid (S_{c_i}, p_{c_i}) \in L_{temp}\} \text{ according to } p_{c_i}$\;
    $\mathcal{A} \leftarrow \mathcal{A} \cup \{S_p \Rightarrow \mathcal{C}_{aligned}\}$\;
}
\Return{$\mathcal{A}$}\;
\end{algorithm}

\subsection{Alignment algorithm}
\label{app:algorithm_details}

To ensure structural traceability and minimize notational noise, Algorithm~\ref{alg:alignment} implements path-coherent alignment by anchoring the SMILES generation of precursors to the corresponding root atoms inherited from the product. Core functions, such as \texttt{RootedSMILES} and \texttt{DefaultRoot}, are implemented using RDKit’s \texttt{rootedAtAtom} and canonical ranking utilities.

\begin{figure}[htb]
    \centering
    \includegraphics[width=0.55\columnwidth]{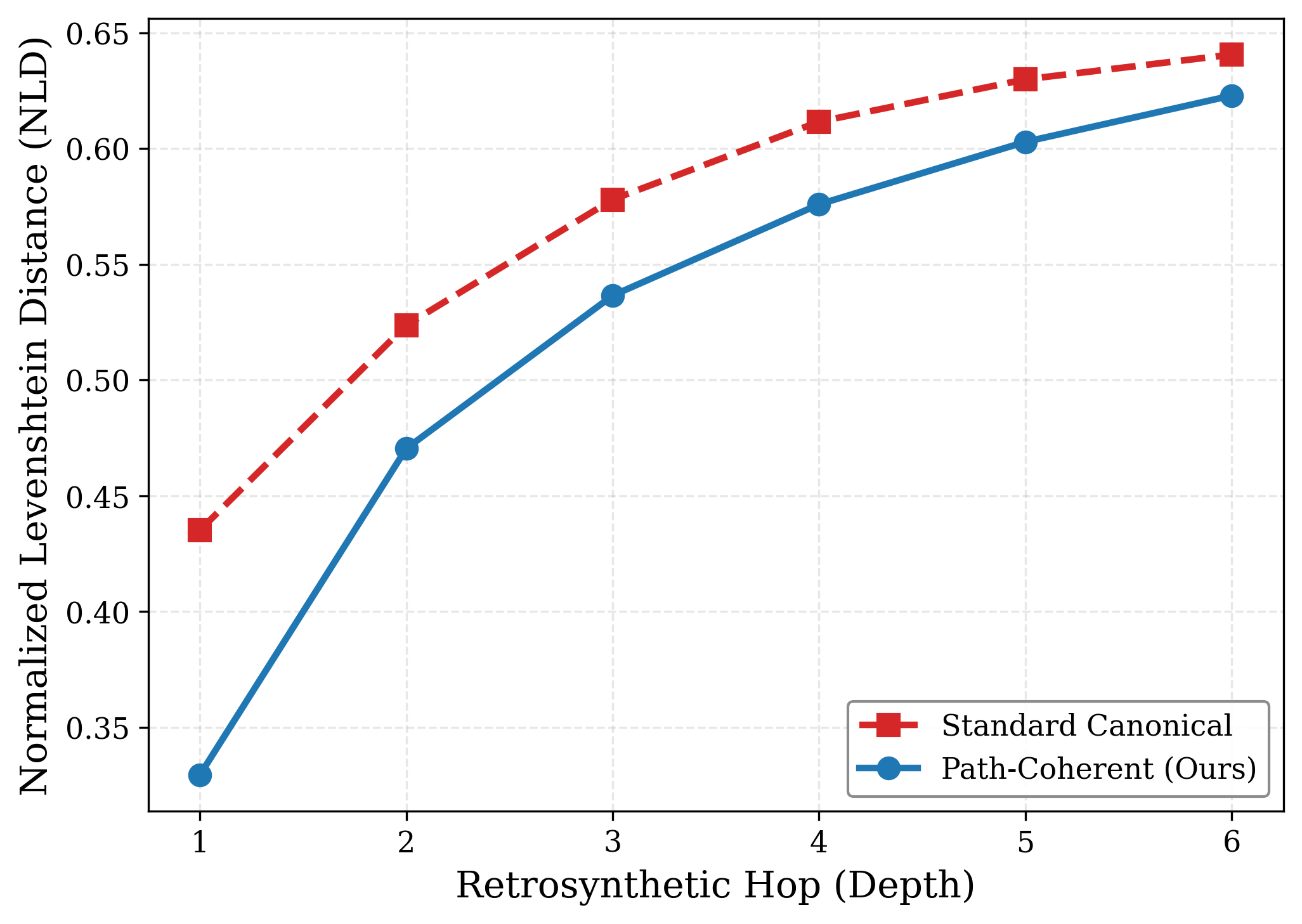}
    \caption{\textbf{Trend of normalized Levenshtein distance (NLD) across retrosynthetic depths.} The consistently lower NLD of our method demonstrates that fragment traceability is preserved throughout the planning horizon, minimizing representational noise.}
    \label{fig:nld_trend}
\end{figure}

\subsection{Quantitative analysis of representational consistency}
\label{app:quantitative_analysis}

We further conduct a quantitative analysis to measure the representational stability of our framework across long-horizon planning trajectories. We employ the Normalized Levenshtein Distance (NLD) to quantify the textual divergence between the target molecule and the generated reaction components.

\textbf{Metric definition.} For a retrosynthetic route, let $S_{\text{target}}$ be the SMILES string of the target molecule, and let $\smash{S_{\text{RHS}}^{(k)}}$ be the SMILES string of the entire right-hand side precursors at step $k$. The NLD at depth $k$ is defined as:
\begin{equation}
    \text{NLD}(k) = \frac{\text{Levenshtein}(S_{\text{target}}, S_{\text{RHS}}^{(k)})}{\max(|S_{\text{target}}|, |S_{\text{RHS}}^{(k)}|)}
\end{equation}

\textbf{Analysis.} General-purpose SMILES notoriously neglects the topological conservation of chemical reactions, creating input-output discrepancies that force models to prioritize syntax over chemical logic \citep{zhong2022root}. While prior works address this locally in single-step predictions, maintaining such stability across multi-step pathways is non-trivial. ReTriP advances this paradigm by introducing a global synchronization mechanism that ensures continuous structural traceability throughout the entire planning horizon.

As illustrated in Figure~\ref{fig:nld_trend}, while the NLD naturally increases with route depth due to cumulative chemical transformations and the shedding of fragments, our path-coherent representation maintains a consistently lower NLD. This advantage remains stable even at extended horizons (Depth $\ge$ 5) where structural divergence is substantial. Consequently, ReTriP successfully decouples genuine chemical transformations from artificial notational drift, freeing the policy from deciphering complex syntactic permutations to focus entirely on strategic structural decomposition.

\subsection{Qualitative comparison}
\label{app:representation_comparison}

To illustrate the impact of our algorithm, we present a side-by-side comparison of a 9-step retrosynthetic route. In \hyperref[box:canonical]{Box~A}, standard canonical SMILES exhibit arbitrary re-indexing, leading to a significant shift in the string head between Step 2 and Step 3. Conversely, our path-coherent representation (\hyperref[box:aligned]{Box~B}) anchors the primary scaffold, ensuring a consistent syntactic prefix across these same steps and throughout the entire planning trajectory.

\phantomsection
\label{box:canonical}
\begin{codebox}{A. Standard canonical representation}
\begin{small}
\textbf{\textsc{Main Path}}
\begin{enumerate}[leftmargin=*, label=\textbf{\arabic*:}, nosep, itemsep=0.5em]
    \item \seqsplit{CN1CCC(c2ccc(Nc3ncc(C(F)(F)F)c(CCc4nccnc4CC(=N)O)n3)cc2)CC1} \\
    $\to$ \seqsplit{NC(=O)Cc1nccnc1CCc1nc(Nc2ccc(C3CCNCC3)cc2)ncc1C(F)(F)F} \\
    \hspace*{0.5em}+ \seqsplit{CC(=O)O[BH-](OC(C)=O)OC(C)=O}
    
    \item \seqsplit{NC(=O)Cc1nccnc1CCc1nc(Nc2ccc(C3CCNCC3)cc2)ncc1C(F)(F)F} \\
    $\to$ \seqsplit{CC(C)(C)OC(=O)N1CCC(c2ccc(Nc3ncc(C(F)(F)F)c(CCc4nccnc4CC(N)=O)n3)cc2)CC1}
    
    \item \seqsplit{CC(C)(C)OC(=O)N1CCC(c2ccc(Nc3ncc(C(F)(F)F)c(CCc4nccnc4CC(N)=O)n3)cc2)CC1} \\
    $\to$ \seqsplit{CCOC(=O)Cc1nccnc1CCc1nc(Nc2ccc(C3CCN(C(=O)OC(C)(C)C)CC3)cc2)ncc1C(F)(F)F} \\
    \hspace*{0.5em}+ \seqsplit{On1nnc2ccccc21}
    
    \item \seqsplit{CCOC(=O)Cc1nccnc1CCc1nc(Nc2ccc(C3CCN(C(=O)OC(C)(C)C)CC3)cc2)ncc1C(F)(F)F} \\
    $\to$ \seqsplit{CCOC(=O)Cc1nccnc1C\#Cc1nc(Nc2ccc(C3CCN(C(=O)OC(C)(C)C)CC3)cc2)ncc1C(F)(F)F}
    
    \item \seqsplit{CCOC(=O)Cc1nccnc1C\#Cc1nc(Nc2ccc(C3CCN(C(=O)OC(C)(C)C)CC3)cc2)ncc1C(F)(F)F} \\
    $\to$ \seqsplit{C\#Cc1nccnc1CC(=O)OCC} \\
    \hspace*{0.5em}+ \seqsplit{CC(C)(C)OC(=O)N1CCC(c2ccc(Nc3ncc(C(F)(F)F)c(Cl)n3)cc2)CC1}
    
    \item \seqsplit{CC(C)(C)OC(=O)N1CCC(c2ccc(Nc3ncc(C(F)(F)F)c(Cl)n3)cc2)CC1} \\
    $\to$ \seqsplit{CC(C)(C)OC(=O)N1CCC(c2ccc(N)cc2)CC1} + \seqsplit{FC(F)(F)c1cnc(Cl)nc1Cl}
    
    \item \seqsplit{CC(C)(C)OC(=O)N1CCC(c2ccc(N)cc2)CC1} \\
    $\to$ \seqsplit{CC(C)(C)OC(=O)OC(C)(C)C} + \seqsplit{O=[N+]([O-])c1ccc(C2CCNCC2)cc1}
\end{enumerate}

\vspace{0.5em}
\hrule
\vspace{0.5em}

\textbf{\textsc{Branch Path}}
\begin{enumerate}[leftmargin=*, label=\textbf{\arabic*:}, nosep, itemsep=0.5em]
    \item \seqsplit{C\#Cc1nccnc1CC(=O)OCC} \\
    $\to$ \seqsplit{CCOC(=O)Cc1nccnc1C\#C[Si](C)(C)C}
    \item \seqsplit{CCOC(=O)Cc1nccnc1C\#C[Si](C)(C)C} \\
    $\to$ \seqsplit{CCOC(=O)Cc1nccnc1Cl} + \seqsplit{C\#C[Si](C)(C)C}
\end{enumerate}
\end{small}
\end{codebox}

\vspace{1em}

\phantomsection
\label{box:aligned}
\begin{codebox}{B. Path-coherent representation (ours)}
\begin{small}
\textbf{\textsc{Main Path}}
\begin{enumerate}[leftmargin=*, label=\textbf{\arabic*:}, nosep, itemsep=0.5em]
    \item \seqsplit{CN1CCC(c2ccc(Nc3ncc(C(F)(F)F)c(CCc4nccnc4CC(=N)O)n3)cc2)CC1} \\
    $\to$ \seqsplit{N1CCC(c2ccc(Nc3ncc(C(F)(F)F)c(CCc4nccnc4CC(N)=O)n3)cc2)CC1} \\
    \hspace*{0.5em}+ \seqsplit{C(C)(=O)O[BH-](OC(C)=O)OC(C)=O}
    
    \item \seqsplit{N1CCC(c2ccc(Nc3ncc(C(F)(F)F)c(CCc4nccnc4CC(N)=O)n3)cc2)CC1} \\
    $\to$ \seqsplit{N1(C(=O)OC(C)(C)C)CCC(c2ccc(Nc3ncc(C(F)(F)F)c(CCc4nccnc4CC(N)=O)n3)cc2)CC1}
    
    \item \seqsplit{N1(C(=O)OC(C)(C)C)CCC(c2ccc(Nc3ncc(C(F)(F)F)c(CCc4nccnc4CC(N)=O)n3)cc2)CC1} \\
    $\to$ \seqsplit{N1(C(=O)OC(C)(C)C)CCC(c2ccc(Nc3ncc(C(F)(F)F)c(CCc4nccnc4CC(=O)OCC)n3)cc2)CC1} \\
    \hspace*{0.5em}+ \seqsplit{n1nn(O)c2ccccc12}
    
    \item \seqsplit{N1(C(=O)OC(C)(C)C)CCC(c2ccc(Nc3ncc(C(F)(F)F)c(CCc4nccnc4CC(=O)OCC)n3)cc2)CC1} \\
    $\to$ \seqsplit{N1(C(=O)OC(C)(C)C)CCC(c2ccc(Nc3ncc(C(F)(F)F)c(C\#Cc4nccnc4CC(=O)OCC)n3)cc2)CC1}
    
    \item \seqsplit{N1(C(=O)OC(C)(C)C)CCC(c2ccc(Nc3ncc(C(F)(F)F)c(C\#Cc4nccnc4CC(=O)OCC)n3)cc2)CC1} \\
    $\to$ \seqsplit{N1(C(=O)OC(C)(C)C)CCC(c2ccc(Nc3ncc(C(F)(F)F)c(Cl)n3)cc2)CC1} \\
    \hspace*{0.5em}+ \seqsplit{C\#Cc1nccnc1CC(=O)OCC}
    
    \item \seqsplit{N1(C(=O)OC(C)(C)C)CCC(c2ccc(Nc3ncc(C(F)(F)F)c(Cl)n3)cc2)CC1} \\
    $\to$ \seqsplit{N1(C(=O)OC(C)(C)C)CCC(c2ccc(N)cc2)CC1} + \seqsplit{c1(Cl)ncc(C(F)(F)F)c(Cl)n1}
    
    \item \seqsplit{N1(C(=O)OC(C)(C)C)CCC(c2ccc(N)cc2)CC1} \\
    $\to$ \seqsplit{N1CCC(c2ccc([N+](=O)[O-])cc2)CC1} + \seqsplit{C(=O)(OC(C)(C)C)OC(C)(C)C}
\end{enumerate}

\vspace{0.5em}
\hrule
\vspace{0.5em}

\textbf{\textsc{Branch Path}}
\begin{enumerate}[leftmargin=*, label=\textbf{\arabic*:}, nosep, itemsep=0.5em]
    \item \seqsplit{C\#Cc1nccnc1CC(=O)OCC} \\
    $\to$ \seqsplit{C(\#Cc1nccnc1CC(=O)OCC)[Si](C)(C)C}
    \item \seqsplit{C(\#Cc1nccnc1CC(=O)OCC)[Si](C)(C)C} \\
    $\to$ \seqsplit{C(\#C)[Si](C)(C)C} + \seqsplit{c1(Cl)nccnc1CC(=O)OCC}
\end{enumerate}
\end{small}
\end{codebox}

\end{document}